\documentclass[iicol,sn-apa]{sn-jnl}

\jyear{2023}%

\theoremstyle{thmstyleone}%
\theoremstyle{thmstyletwo}%

\theoremstyle{thmstylethree}%
\usepackage{array,multirow,graphicx,adjustbox}
\usepackage{booktabs}  
\usepackage{multirow}
\usepackage{pbox}
\usepackage{amsmath}

\usepackage{amsmath,amsfonts,bm}

\def\eqref#1{equation~\ref{#1}}

\def\1{\bm{1}}

\def\vb{{\bm{b}}}

\def\mC{{\bm{C}}}

\def\mK{{\bm{K}}}

\def\mP{{\bm{P}}}

\def\mV{{\bm{V}}}
\def\mW{{\bm{W}}}

\def\mZ{{\bm{Z}}}

\DeclareMathAlphabet{\mathsfit}{\encodingdefault}{\sfdefault}{m}{sl}
\SetMathAlphabet{\mathsfit}{bold}{\encodingdefault}{\sfdefault}{bx}{n}
\newcommand{\tens}[1]{\bm{\mathsfit{#1}}}

\def\tB{{\tens{B}}}

\def\tK{{\tens{K}}}

\def\tQ{{\tens{Q}}}

\def\tV{{\tens{V}}}

\usepackage{hyperref}
\usepackage{url}

\usepackage{color, colortbl}
\definecolor{grayrow}{cmyk}{0,0,0,0.18}
\definecolor{Image}{RGB}{93,59,155}
\definecolor{Video}{RGB}{105,145,60}

\usepackage{arydshln}
\begin{document}
\title[Article Title]{Towards a Unified View on Visual Parameter-Efficient Transfer Learning}

\author[1]{\fnm{Bruce X.B.} \sur{Yu}}\email{bruce.xb.yu@connect.polyu.hk}

\author[2]{\fnm{Jianlong} \sur{Chang}}\email{jianlong.chang@huawei.com}

\author[1]{\fnm{Lingbo} \sur{Liu}}\email{lingbo.liu@polyu.edu.hk}

\author[2]{\fnm{Qi} \sur{Tian}}\email{tian.qi1@huawei.com}

\author*[1]{\fnm{Chang Wen} \sur{Chen}}\email{changwen.chen@polyu.edu.hk}

\affil*[1]{\orgdiv{Department of Computing}, \orgname{The Hong Kong Polytechnic University}, \orgaddress{ \city{Hong Kong}, \postcode{999077},  \country{China}}}

\affil[2]{ \orgname{Huawei Inc.}, \orgaddress{ \city{Shenzhen}, \postcode{518000}, , \country{China}}}

\abstract{
Parameter efficient transfer learning (PETL) aims at making good use of the representation knowledge in the pre-trained large models by fine-tuning a small number of parameters. Recently, taking inspiration from the natural language processing (NLP) domain, popular PETL techniques such as prompt-tuning and Adapter have also been successfully applied to the vision domain. However, prefix-tuning remains under-explored for vision tasks. In this work, we intend to adapt large vision models (LVMs) to downstream tasks with a good parameter-accuracy trade-off. Towards this goal, we propose a framework with a unified view of PETL called visual-PETL (V-PETL) to investigate the effects of different PETL techniques, data scales of downstream domains, positions of trainable parameters, and other aspects affecting the trade-off. 
Specifically, we analyze the positional importance of trainable parameters and differences between NLP and vision tasks in terms of data structures and pre-training mechanisms while implementing various PETL techniques, especially for the under-explored prefix-tuning technique. Based on a comprehensive understanding of the differences between NLP and vision data, we propose a new variation of the prefix-tuning module called parallel attention (PATT) for vision downstream tasks. 
An extensive empirical analysis on vision tasks via different frozen LVMs has been carried and the findings show that the proposed PATT can effectively contribute to other PETL techniques. An effective scheme Swin-BAPAT derived from the proposed V-PETL framework achieves significantly better performance than the state-of-the-art AdaptFormer-Swin with slightly more parameters and outperforms full-tuning with far fewer parameters. 
Code and data are available at: \url{https://github.com/bruceyo/V-PETL}}.




\maketitle

\section{Introduction}
\begin{figure}
\begin{center}
\includegraphics[width=\linewidth]{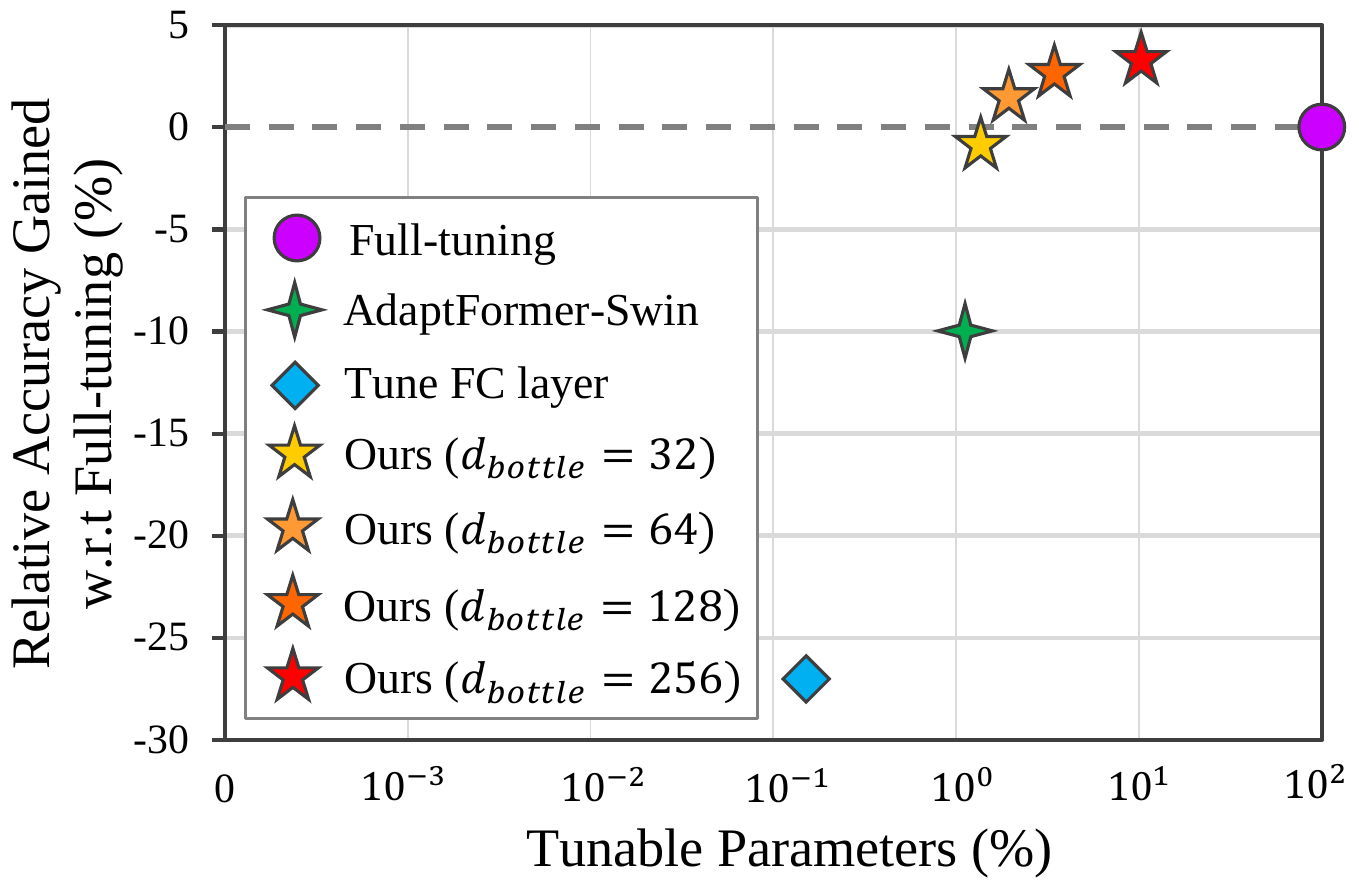}
\end{center}
\caption{Parameter-accuracy trade-off. Adapting backbone Swin-B \citep{liu2022video} pre-trained on Kinetics 400 via different fine-tuning methods on the something-something v2 \citep{goyal2017something} dataset. Our methods perform significantly better than the state-of-the-art AdaptFormer-Swin \citep{chen2022adaptformer} (our implementation with batch size $16$) with slightly more tunable parameters and outperform full-tuning with increasing margins when using larger values of $d_{bottle}$.}
\label{fig:performance}
\end{figure}

Many vision tasks rely on fine-tuning pre-trained large vision models (LVMs) to achieve good performance. One standard modus operandi of transfer learning consists of two steps: pre-train a model on a source domain and fine-tune the entire model on a target domain \citep{zhuang2020comprehensive}. 
Despite that prior works have achieved promising performance, such vanilla practice of fine-tuning is faced with challenges for adopting LVMs to downstream tasks. First, this full-tuning strategy requires one to update and store separate model parameters for different downstream tasks, which can be expensive and infeasible for the era of increasingly large vision models from EfficientNet-based \citep{pham2021meta} ($480$M parameters) to Transformer-based \citep{yu2022coca} ($2,100$M parameters) ones. Recently, the vision model size has even been scaled to $22$B parameters \cite{dehghani2023scaling}, leading to increased performance for downstream tasks by linear probing (i.e. freeze the pre-trained model). For such large vision models, making good use of shared parameter weights deployed on the cloud can be beneficial for edge devices such as autonomous vehicles, and drones who are intensive in computing and battery resources \citep{yuan2022roadmap}.
Second, the full fine-tuning strategy relies on high-quality downstream data and can hardly adapt to unseen scenarios that have large distribution shift \citep{kumar2021fine}, which is unlike the learning process of humans who can learn from few samples and generalize well to new circumstances. This issue has been researched in directions such as zero-shot learning, few-shot learning, and continual learning \citep{li2021else}. Another popular strategy is fine-tuning the downstream task head, i.e., the last fully connected (FC) layer, to avoid tuning the whole backbone model, which usually leads to poor performance when the target domain is large in data scale (see Figure \ref{fig:performance}). 
Given the paradigm of fine-tuning increasingly large vision models, how to transfer such LVMs with parameter-accuracy trade-off is a hot topic in various domains \citep{ijcai2022p769,sung2022vl,lin2020exploring,houlsby2019parameter}. 

Taking the video-based action recognition task as an example, it can be inconvenient for deploying increasingly large vision models to edge devices such as autonomous driving \citep{liu2019edge} and unmanned aerial vehicle \citep{li2021uav} as they can heavily rely on the interaction with cloud services for adapting to new environments via active learning  \citep{wang2021interactive} or continual learning \citep{li2021else}. Re-training LVMs on the cloud is usually not cost-effective due to the expensive overheads of storage and computational resources. Furthermore, these resources are limited on edge devices such as autonomous vehicles and unmanned aerial vehicles, making the sense for developing effective fine-tuning methods with proper parameter-accuracy trade-offs that can be fine-tuned on edge devices and interacting with the LVMs deployed on the cloud. 

 We note that fine-tuning visual-language pre-trained (VLP) models \citep{gan2022vision} identified with key significance regarding capability and homogenization can be a promising direction \cite{bommasani2021opportunities}. As such, a couple of recent works \cite{zang2022unified, pan2022st} achieved promising performance on vision downstream tasks by finetuning VLP models such as CLIP \citep{radford2021learning} and ALIGN \citep{jia2021scaling}, Florence \citep{yuan2021florence}, BEiT \citep{wang2022image}, GATO \citep{reed2022generalist}, CoCa \citep{yu2022coca}, SWAG \citep{singh2022revisiting} etc. 
 However, according to the results in \citep{pan2022st} and \citep{he2022parameter}, fine-tuning VLP models do not lead to results as good as fine-tuning  supervised pre-trained vision models. In addition, purely vision models are also increasingly large (reach 22B parameters) \citep{dehghani2023scaling} and gain great advances recently \citep{chen2020generative,li2021efficient,reed2022generalist} with varied pre-training strategies \citep{khan2022transformers, zhou2023comprehensive}, so we focus on fine-tuning pre-trained pure vision models.

There have been some pioneering works for the PETL of visual models such as AdaptFormer \citep{chen2022adaptformer} and visual prompt tuning (VPT) \citep{jia2022visual}.  AdaptFormer is primarily proposed based on vision transformer \citep{zhai2022scaling}, representing one of the state-of-the-art large models for image-based tasks. The proposed Adapter module directly brings from \cite{houlsby2019parameter} due to its convenience of being inserted into any model. Implementing with a large batch size of $1,024$ with $64$ GPUs, Adaptformer shows a promising parameter-accuracy trade-off on video data. However, such powerful computing resource is not realistic for the usage of edge devices. Meanwhile, whether the good trade-off can be maintained for small batch size remains under-explored. Inspired by the Prompting in NLP \citep{liu2021pre}, VPT proposes visual-prompt to fine-tune visual models for image-based tasks. According to the empirical results in \cite{chen2022adaptformer}, Adapter modules achieve superior performance over VPT in the regimes of both self-supervised and supervised pre-training. Another concern of VPT is its insertion of trainable parameters to LVMs can be similar to prefix-tuning. Hence, we do not continue to compare our method with VPT but investigate the effect of prefix-tuning and mainly compare our proposed method with the Adapter on varied vision downstream tasks.

Taking the recent inspiration of the mix-and-match Adapter (MAM Adapter) \citep{he2022towards} in the NLP domain, we aim to propose a unified model for the vision domain, especially for video-based downstream tasks. \cite{he2022towards} analyzed the unified view among PETL techniques such as prefix-tuning, low-rank (LoRA) adaptation, and Adapter, pointing out the similarity between prefix-tuning and Adapter in terms of calculating the attention. The difference is that the former performs weighted addition while the latter ones are unweighted. It is worth noting that prefix-tuning has not ever been applied to visual tasks in the form of pure vision models due to the intrinsic differences regarding pre-training methods of NLP and vision models. Another obstacle to directly applying prefix-tuning to visual tasks is the structural difference between text and vision data (we further discuss this in Section \ref{subsec:connection}). Considering the characteristics of vision data, we propose a new variation of the prefix-tuning module called parallel attention (PATT) to adapt LVMs to different downstream vision tasks. The differences between our method comparing the original prefix-tuning in NLP are twofold: prefix calculation and the manner of insertion (see Figure \ref{fig:petls}[b] and Figure \ref{fig:petl_patt}). Regarding the implemented PVMs, we focus on Video Swin Transformer \citep{liu2022video}, one of the state-of-the-art vision models that bring competitive performance on large-scale action recognition datasets such as Kinetics 400 and 600 \cite{kay2017kinetics}. Extensive experiments have also been conducted to verify the effectiveness of our PATT on more vision downstream tasks (e.g., image classification) with different cross-domain settings (i.e., fine-tuning models pre-trained on image tasks to video tasks) and more PVMs such as supervised and self-supervised ViTs \citep{he2022masked, tong2022videomae}.

The main contributions of this paper can be threefold as follows:\\
1. We analyze different PETL techniques using the backbone model (i.e., LVM) Swin Video Transformer for video-based tasks, providing a unified view via our V-PETL framework and investigating the importance of the fine-tuning position. \\
2. Based on the comprehensive understanding of intrinsic differences between NLP and vision data regarding data structures and pre-training mechanisms, we leverage prefix-tuning to our V-PETL with a new variation called PATT.\\
3. Upon extensive ablation experiments regarding various effect factors, we empirically validate the promising parameter-accuracy trade-off achieved by our adjustable and easy-to-use PATT module, contributing to the existing literature of PETL techniques for making good use of future LVMs.

\section{Unified Framework}
\subsection{Recap of PETL Techniques}\label{subsec:recap}

\begin{figure*}[h]
\begin{center}
\includegraphics[width=\linewidth]{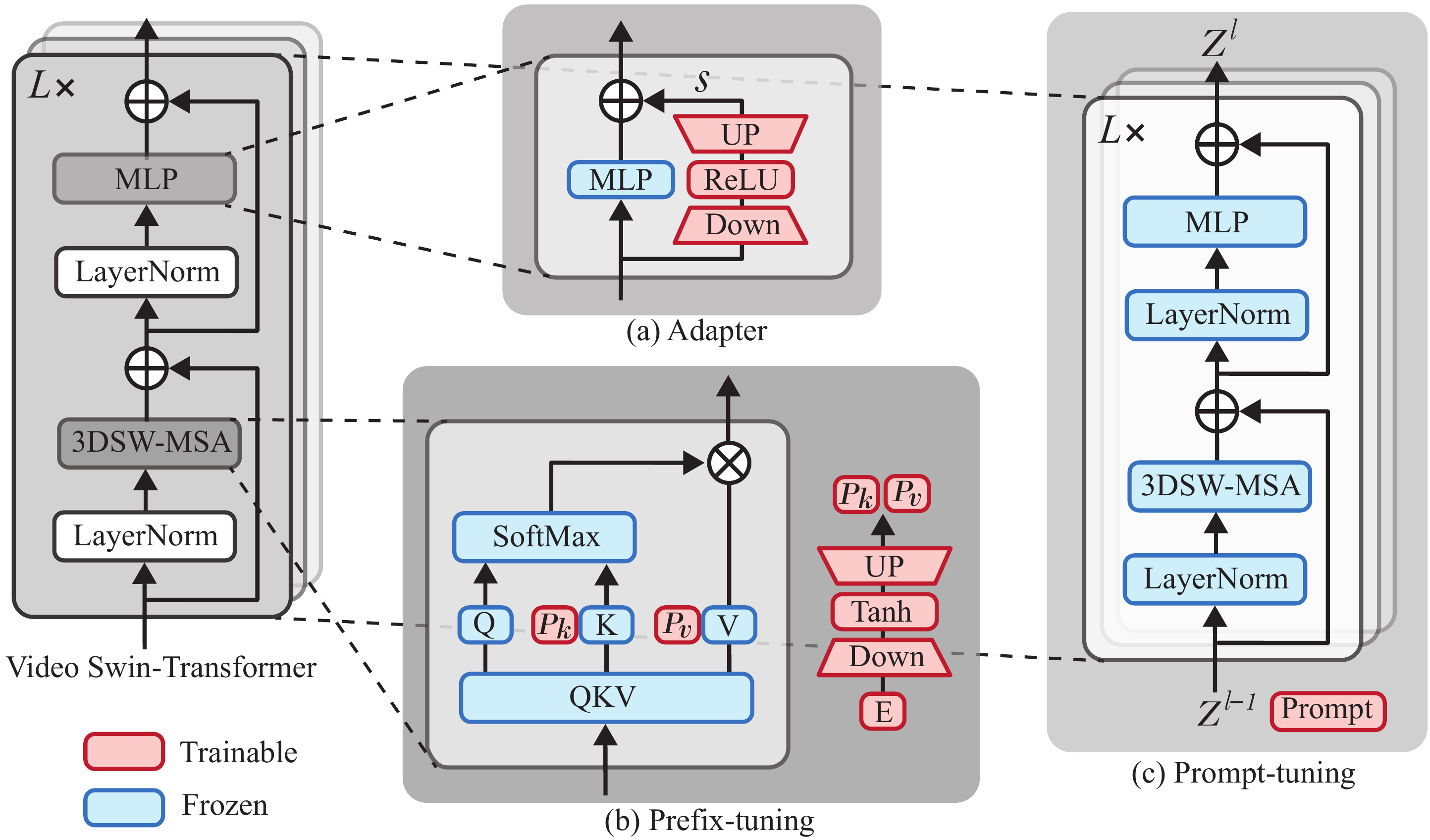}
\end{center}
\caption{V-PETL: A unified view of visual PETL techniques. They bring trainable parameters to different positions of the backbone model in various manners. AdaptFormer and prefix-tuning respectively perform at the MLP and 3DSW-MSA modules that can adjust the number of trainable parameters via the bottleneck size of down and up projections. While prompt-tuning performed at the layer level can adjust the length of prompts to control the tuned parameters.}
\label{fig:petls}
\end{figure*}

\subsection{Recap of Video Swin Transformer}
Video Swin Transformer \citep{liu2022video} is formed with Transformer layers (a.k.a. stages) that are consisted with 3D Video Swin Transformer blocks. With varied layers, blocks, and channel sizes, the model can be formed as Swin-T, Swin-S, Swin-B, and Swin-L. The basic architecture of a 3D Swin Transformer block is shown in Figure \ref{fig:petls}, which is mainly composed of a 3D shifted window-based multi-head self-attention (3DSW-MSA) module and a fully connected feed-forward network (FFN) implemented with a 2-layer MLP. Layer normalization (LN) and residual connection are respectively performed before and after both FFN and 3DSW-MSA modules. One such Video Swin Transformer block can be represented as:
\begin{equation}
\begin{split}
  \hat{\displaystyle \mZ}^l&=\text{3DSW}{\text -}\text{MSA}(\text{LN}(\displaystyle \mZ^{l-1}))+\displaystyle \mZ^{l-1},\\
  \displaystyle \mZ^{l}&=\text{FFN}(\text{LN}(\hat{\displaystyle \mZ}^l))+\hat{\displaystyle \mZ}^l,
 \end{split}
 \label{eq:vit-swin}
\end{equation}
where $\hat{\displaystyle \mZ}^l$ and $\displaystyle \mZ^{l}$ respectively indicate the output of 3DSW-MSA and FNN modules. 

Given a video input sized $t\times w\times h\times 3$, containing $t$ video frames with their heights and widths being $h$ and $w$, respectively. The 3D patch for video data sized $2\times 4\times 4\times 3$ is treated as a token. Then we will have $\frac{t}{2}\times \frac{w}{4}\times \frac{h}{4}$ 3D tokens after a 3D patch partitioning layer. 
Given the 3D tokens sized $\frac{t}{2}\times \frac{w}{4}\times \frac{h}{4}$ and a 3D window with the size of $p\times m\times m$, the self-attention module, using the regular window partition strategy, will partition the 3D tokens to $\frac{t}{2p}\times \frac{w}{4m}\times \frac{h}{4m}$ non-overlapping windows. For shifted 3D window, the partition is shifted along the temporal, height, and width dimensions by $\frac{p}{2}\times \frac{m}{2}\times \frac{m}{2}$. For example, if we have an input video sized $8\times 224\times 224\times 3$ and a $8\times 7\times 7$ 3D window, after the patch embedding, we will have $4\times 56\times 56$ 3D tokens with each of them sized $2\times 4\times 4\times 3$. Without shifting, the non-overlapping window size will be $1\times 8\times 8 = 64$.Then through the 3D window shifted by $(4, 3, 3)$ , the number of 3D windows becomes $1\times 9\times 9 = 81$.

The 3DSW-MSA module is formed with a 3D relative position bias $\displaystyle \tB \in \mathbb{R}^{p^2{\times}m^2{\times}m^2}$, each of which can be represented as:
\begin{equation}
Attention(\displaystyle \tQ,\displaystyle \tK,\displaystyle \tV) = SoftMax(\frac{\displaystyle \tQ\displaystyle \tK^T}{\sqrt{d}} + \displaystyle \tB)\displaystyle \tV,
  \label{eq:att}
\end{equation}
where $\displaystyle \tQ,\displaystyle \tK,\displaystyle \tV \in \mathbb{R}^{p\times m{\times}m\times d}$ are the query, key, and value matrices, $p\times m{\times}m$ is the number of tokens and $\displaystyle d$ is the dimension of the tokens. MSA simultaneously performs the attention mechanism for $n_{head}$ heads, where the $i$th head can be parameterized by $\displaystyle \mW_q^{(i)},\displaystyle \mW_k^{(i)},\displaystyle \mW_v^{(i)} \in \mathbb{R}^{d \times 3d}$,  projecting the input $Z^{l-1}$ to queries, keys, and values. Given a matrix $\displaystyle \mC \in \mathbb{R}^{\widetilde{m}\times d}$, $\widetilde{m}=p\times m{\times}m$, for performing attention, the 3DSW-MSA can be calculated as:
\begin{equation}
\begin{split}
\text{3DSW}{\text -}\text{MSA}(\displaystyle \mZ^{l-1},\displaystyle \mC)= \\
Concat(head_1,...,head_n)\displaystyle \mW_o,\\
head_i=\\
Attention(\displaystyle \mZ^{l-1}\displaystyle \mW_q^{(i)},\displaystyle \mC\displaystyle \mW_k^{(i)},\displaystyle \mC\displaystyle \mW_v^{(i)}),
\end{split}
\label{eq:3DSW}
\end{equation}

where $\displaystyle \mW_o$ is the parameters of a linear project layer.
The FNN module is composed of two linear layers with a  GELU activation function in between, which can be computed as:
\begin{equation}
  \text{FFN}(\hat{\displaystyle \mZ}^l) = \text{GELU}(\text{LN}(\hat{\displaystyle \mZ}^l)\displaystyle \mW_1+\displaystyle \vb_1)\displaystyle \mW_2 + \displaystyle \vb_2,
  \label{eq:ffn}
\end{equation}
where $\displaystyle \mW_1\in \mathbb{R}^{d_{hidden}{\times}d}$, $\displaystyle \mW_2 \in \mathbb{R}^{d{\times}d_{hidden}}$, $\displaystyle \vb_1 \in \mathbb{R}^{d_{hidden}}$, and  $\displaystyle \vb_2 \in \mathbb{R}^{d}$. The value of $d_{hidden}$ usually takes a large value (e.g., $d_{hidden}=4d$).

\textbf{Prefix-tuning \citep{li2021prefix}}:
The prefix-tuning approach prepends learnable prefix tokens to the keys and values of the MSA module of the model (see Figure \ref{fig:petls}[b]). Specifically, two prefix matrices $\displaystyle \mP_k, \displaystyle \mP_v \in \mathbb{R}^{d_{token}{\times}d}$ that are randomly initialized with $d_{token}$ tokens and transformed from two linear layers (with parameters $\displaystyle \mW_{pk}^{(i)} \in \mathbb{R}^{d{\times}d_{middle}}$ and $\displaystyle \mW_{pv}^{(i)} \in \mathbb{R}^{d_{middle}{\times}d}$) and a Tanh layer in between are concatenated to the original key and value, leading the calculation of $head_i$ in Eq. \ref{eq:3DSW} to:
\begin{equation}
\begin{split}
head_i=Attention(\displaystyle \mZ^{l-1}\displaystyle \mW_q^{(i)}, 
\\
concat(\displaystyle \mP_k^{(i)}, \displaystyle \mC\displaystyle \mW_k^{(i)}),
\\
 concat(\displaystyle \mP_v^{(i)}, \displaystyle \mC\displaystyle \mW_v^{(i)})),
\end{split}
  \label{eq:head}
\end{equation}
where the $concat$ is the concatenation performed along the token dimension to mimic the prefix-tuning in NLP tasks. Here, a question regarding whether this direct implementation will work for the vision domain is raised (results are in Table \ref{tab:comparison_petls_qkv_combinations}). This direct implementation is empirically invalid and we make further modifications to it in Section\ref{subsec:connection}.

\textbf{Adapter \citep{chen2022adaptformer}}:
Inspired by the works of \cite{houlsby2019parameter,he2022towards} for PETL in NLP tasks, Adapter \citep{chen2022adaptformer} has been directly used for vision tasks, showing promising performance using far less tunable parameters. The number of parameters of Adapter is controlled by a parameter $d_{bottle}$ $(d_{bottle}\ll d)$, adjusting the space size of a low-dimensional representation. The Adapter module first uses a down-projection with $\displaystyle \mW_{down} \in \mathbb{R}^{d{\times}d_{bottle}}$ to project the feature to the lower-dimensional representation, followed by a ReLU activation function, and an up-projection with $\displaystyle \mW_{up} \in \mathbb{R}^{d_{bottle}{\times}d}$. 
\begin{equation}
  \widetilde{\displaystyle \mZ}^{l}=\text{ReLU}(\text{LN}(\hat{\displaystyle \mZ}^l)\displaystyle \mW_{down})\displaystyle \mW_{up},
  \label{eq:adapter_z_wave}
\end{equation}
then two positions implementing Adapter (parallel and sequential) can be respectively computed as:
\begin{equation}
\begin{split}
    \displaystyle \mZ^{l}&=\text{FFN}(\text{LN}(\hat{\displaystyle \mZ}^l))+\hat{\displaystyle \mZ}^l + s\widetilde{\displaystyle \mZ}^{l},\ and\\
  s\displaystyle \mZ^{l}&=\text{ReLU}(\text{FFN}(\text{LN}(\hat{\displaystyle \mZ}^l))\displaystyle \mW_{down})\displaystyle \mW_{up}+\hat{\displaystyle \mZ}^l,
\end{split}
\end{equation}
where $s$ is a scalar, controlling the effect of the Adapter (will be ablated in experiments). According to  \cite{chen2022adaptformer}, the parallel implementation (see Figure \ref{fig:petls}[a]) empirically performs better. 

\textbf{Prompt-tuning \citep{jia2022visual}}:
Prompt-tuning (see Figure \ref{fig:petls}[c]) is inspired by the success of prompt-tuning that adapts large-scale models to varied downstream NLP tasks.
The idea of VPT \citep{jia2022visual} is to fine-tune a learnable matrix  $\displaystyle \mP_{prompt}^{l-1} \in \mathbb{R}^{d_{prompt}{\times}d}$, $d_{prompt} < d_{token}-1$ for the $l$th Transformer layer or all Transformer layers, which are known as shallow prompt and deep prompt, respectively. 
\begin{equation}
\begin{split}
  \hat{\displaystyle \mZ}^l=\text{3DSW}{\text -}\text{MSA}(\text{LN}([x^{l-1}, \\
  \displaystyle \mP_{prompt}^{l-1},\displaystyle \mZ^{l-1}]))+\displaystyle \mZ^{l-1},
\end{split}
\end{equation}
where $x^{l-1} \in \mathbb{R}^{d}$ denotes the [CLS]'s embedding for the $l$th layer's input space, $\displaystyle \mP_{prompt}^{l-1}$ is implemented by overlapping the top $d_{prompt}$ tokens of $\displaystyle \mZ^{l-1}$ \citep{jia2022visual}. While it has also been implemented in front of the $x^{l-1}$ \citep{chen2022adaptformer}.

\begin{figure*}[h]
\begin{center}
\includegraphics[width=0.93\linewidth]{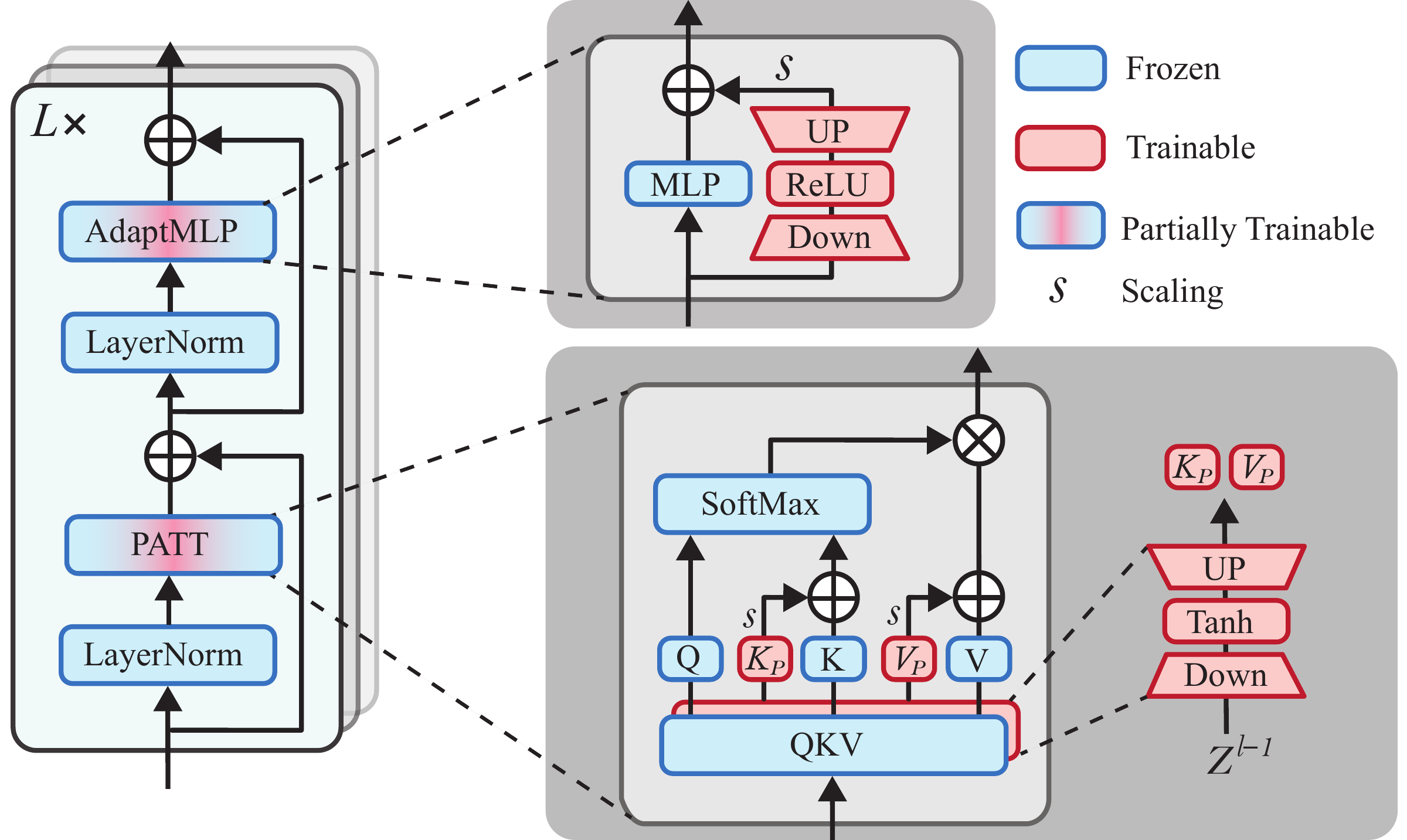}
\end{center}
\caption{Structure of Swin-BAPAT, including partially trainable modules: the proposed PATT and AdaptMLP. For PATT, red parts are trainable parameters calculated by the same input for preparing query, key, and value (i.e., the output of the previous layer passing through a layer normalization layer $\displaystyle \mZ^{l-1}$).}
\label{fig:petl_patt}
\end{figure*}

\textbf{Others}:
Other PETL techniques include ST-Adapter \citep{pan2022st}, LoRA \citep{hu2022lora}, BitFit \citep{zaken2022bitfit}, KAdaptation \citep{he2022parameter}, etc. ST-Adapter adapts image-text models pre-trained on large scale datasets such as 400M image-text pair proposed by CLIP \citep{radford2021learning} and the IG-3.6B used by SWAG \citep{singh2022revisiting} to video understanding downstream tasks, which matches and even outperforms full-tuning. LoRA approximates the optimization process by injecting learnable low-rank matrices into the attention module. This method does not show superior performance for NLP tasks in terms of parameter efficiency. Hence, we do not prioritize this direction in this work. BitFit only tunes the bias terms of the backbone models, making it very parameter-efficient. KAdaptation aims to learn a shared-weight matrix to automatically choose the submodule of the pre-trained large model via the Kronecker product, which can be regarded as a weight-sharing neural architecture search method \citep{xie2021weight}.

\subsection{Revisiting Prefix-tuning for Visual Tasks}\label{subsec:connection}
The prefix implementation in NLP \cite{li2021prefix,he2022towards} can be regarded as prepending contextual information for downstream tasks, which is similar to the pre-training process aiming to predict masked words in the process of an inner loop \citep{brown2020language}. The learning process of prefix-tuning can also be regarded as similar to the learnable prompt of Prompt-tuning \citep{jia2022visual}. Prompt tokens embedded from text templates can help adapt downstream tasks to a frozen large-scale pre-trained NLP model. However, considering the pre-training process of pure vision models, such a direct implementation might not make sense for visual tasks. 

Although autoregressive pre-training has been conducted in the visual domain \citep{he2022masked,tong2022videomae}, adding a prefix for a sentence input in NLP can be structurally different from the visual domain. Specifically, masked pixels in an image or video data cannot be regarded as some word-level semantic information (e.g., cloze or prefix) as in the NLP domain. As such, there remains a lack of proper interpretation for what kind of prefix or prompt can be learned for pure vision models. Especially for cross-domain adaptation, vision models pre-trained with a single task might have no knowledge foundation that enables downstream tasks to adapt to them, making fine-tuning self-supervised models usually underperform fine-tuning supervised models \citep{kim2022broad}. 

Recall that the embedding state of prefix-tuning is randomly initiated, which is known as learnable prefix but can bring random noise that later turns out to affect the convergence of the fine-tuning downstream tasks. Hence, inspired by the connection between Adapter and prefix-tuning \citep{he2022towards}, we avoid such learnable prefix design with random initialization and propose a parallel attention (PATT) to the original attention module (see Figure \ref{fig:petl_patt}).  It is worth noting that our PATT is different with the modification of prefix-tuning called Multi-head Parallel Adapter (MH PA) in \cite{he2022towards}. MH PA intrinsically added an Adapter to the output of the whole attention module, which brings a similar effect to adding an Adapter at the MLP module according to the results in \cite{he2022towards}. Whereas our method can be regarded as a more straightforward modification of prefix-tuning as it targets projection layers of key and value, which leads to more possible further variants (see Table \ref{tab:comparison_petls_qkv_combinations}). The Adapter structure can effectively control the number of trainable parameters via $d_{bottle}$, which is similar to the effect of the middle dimension $d_{middle}$ of $\displaystyle \mW_{pk}^{(i)}$ and $\displaystyle \mW_{pv}^{(i)}$ 
for preparing the prefix. Specifically, for the $l$th layer, we use output of its previous layer $Z^{l-1}$ and project it to a pair of matrices $\displaystyle \mK_p,\displaystyle \mV_p \in \mathbb{R}^{\widetilde{m}\times d}$ via a similar mechanism of Eq. \ref{eq:adapter_z_wave}:
\begin{equation}
  \displaystyle \mK_p, \displaystyle \mV_p=\text{Tanh}(\displaystyle \mZ^{l-1}\displaystyle \mW_{down})\displaystyle \mW_{up},
  \label{eq:prefix_p_wave}
\end{equation}

where $\text{Tanh}$ is the activation function used for preparing the prefix, which can be replaced by other activation functions such as $\text{RELU}$ and $\text{GELU}$. Here, we follow the original prefix implementation as its value ranges from $- 1$ to $1$. Given $\displaystyle \mK_p$ and $\displaystyle \mV_p$, Eq. \ref{eq:head} can be rewritten as:
\begin{equation}
\begin{split}
head_i=Attention(\displaystyle \mZ^{l-1}\displaystyle \mW_q^{(i)}, \\
s\displaystyle \mK_p+\displaystyle \mC\displaystyle \mW_k^{(i)}, s\displaystyle \mV_p+ \displaystyle \mC\displaystyle \mW_v^{(i)}),
  \label{eq:head2}
  \end{split}
\end{equation}
where $s$ is a scalar for adjusting the effect of PATT. Note that without considering the physical meaning of such design, for PETL purpose, one can perform similar practice for any combinations of $\displaystyle \tQ$, $\displaystyle \tK$, and $\displaystyle \tV$. This brings connection to the LoRA \citep{hu2022lora} method, which add parallel trainable parameters to $\displaystyle \tQ$ and $\displaystyle \tV$. Empirically, where to perform the PATT makes little difference, but the amount of  trainable parameters brings larger effect for large-scale downstream domains.

\subsection{V-PETL: Unified View on Visual PETL}\label{subsec:v_petl}

\begin{table}
    \caption{Comparison of independently fine-tuning varied positions of the video swin transformer block on SSv2.}
  \label{tab:position}
\begin{center}
    \resizebox{\linewidth}{!}{%
  \begin{tabular}{lcc}
    \toprule
  \multicolumn{1}{l}{\bf Position}  &\multicolumn{1}{c}{\bf \# Params} &\multicolumn{1}{c}{\bf Top-1 (\%)} \\
  \midrule
      Full-tuning   &  87.82M  & 50.99 \\
      Tune FC Layer &  0.18M  & 24.13 \\
        \midrule
    LayerNorm 1  &  0.02M  & 14.35 \\
    Attn, Proj  &  6.99M  & 47.58 \\
    Attn, QKV  &  20.98M  & 50.02 \\
    Attn, SoftMax  &  0.95M  & 27.67 \\
    LayerNorm 2  &  0.02M  & 14.62 \\
    MLP, FC1  &  27.97M  & 47.10 \\
    MLP, FC2  &  27.93M  & 45.32 \\
    DownSample  &  2.76M  & 27.53 \\
    \bottomrule
  \end{tabular}

  }
  \end{center}
\end{table}
Given the PETL techniques at hand, there can be many potential combinations leading to good parameter-accuracy trade-offs. However, it is unrealistic to  exhaustively test all the methods for a specific downstream task. Other than probing such solutions via evolutionary search as in \cite{zhang2022neural}, we aim to propose more understandable models by empirically analyzing the effect of different designs independently. According to the preliminary results shown in Figure \ref{fig:performance}, we argue that the position and amount of parameters are important for PETL techniques, especially when the target domain is not small. 

\begin{table*}[t]
\renewcommand{\arraystretch}{1.2}
    \centering
    \caption{Fine-tuning settings of downstream tasks. The upper part shows  shared configurations such as optimizer, base learning rate, etc.; the lower part shows the separated ones.}
    \label{tab:settings}
\begin{tabular}{lcc}
\hline
Configuration & Video & Image  \\
\hline
Optimizer & \multicolumn{2}{c}{SGD}  \\
Base learning rate & \multicolumn{2}{c}{0.1} \\
Learning rate schedule & \multicolumn{2}{c}{Cosine decay}
\\
Optimizer momentum & \multicolumn{2}{c}{0.9} \\
GPU number    & \multicolumn{2}{c}{4}  \\
 \hdashline
Batch size & 64 & 256  \\
Warm-up epochs  & 10 & 20  \\
Training epochs & 70 & 100  \\
Augmentation   & RandomResizedCrop & MultiScaleCrop \\
\hline
\end{tabular}
\end{table*}

To verify the importance of position and tuned parameter amount, we independently tune different modules of the backbone model. Table \ref{tab:position} shows the results. We can see that the attention module's QKV layer has $20.98$M parameters while the MLP module has the most number of parameters of $55.90$M. Tuning positions with more parameters will lead to better performance for SSv2. Thanks to the bottleneck mechanism of Adapter and prefix-tuning, one can effectively achieve a good parameter-accuracy trade-off. As such, we derive a model called Swin-B-Adapter-PATT (Swin-BAPAT) from the V-PETL framework by using the parallel Adapter and our PATT to leverage the adaption of pre-trained backbone model at the positions of attention and MLP modules, respectively. Fig. \ref{fig:petl_patt} demonstrates the structure of the proposed Swin-BAPAT with its corresponding trainable parameters at MLP and Attention modules of the Swin Video Transformer model. In addition to the Adapter and PATT, we also fine-tune the last fully connected layer as it has a relatively smaller amount of tunable parameters (i.e, $0.18$M) than the Adapter and PATT.

\section{Experiments}
\subsection{Experimental Settings}
\subsubsection{Downstream Tasks}
We mainly evaluate our PATT module on the video-based human action recognition task to verify its effectiveness. In addition, we also expand our method to image-based tasks. 

\textbf{Video-based tasks}:Something-something v2 (SSv2 \citep{goyal2017something}) It has 108,499 short videos for $174$ human-object interaction categories with durations between $2$ to $6$ seconds. The challenge of this dataset is that it contains $23,137$ distinct object names with an imbalanced distribution. The original dataset is split into train, validation, and test sets with a ratio of 8:1:1. The extended version (SSv2) of this dataset is consisted of $168,913$ training samples, $24,777$ validation samples, and $27,157$ testing samples with the sample number of action labels. The training and testing samples are used. HMDB51 \citep{kuehne2011hmdb} contains $6,766$ video samples for $51$ action categories including videos of varied visible body parts, camera motion, camera view, and clip quality. All video samples have at least $101$ clips and a minimum height of $60$ pixels for actors. The original dataset has three splits of training and evaluation. We follow existing work \cite{chen2022adaptformer} by using the first training and evaluation split that has $3,570$ and $1,530$ samples, respectively. 

\textbf{Image-based tasks}: Following the experimental set ups in AdaptFormer \citep{chen2022adaptformer}, three datasets: CIFAIR-100 \cite{krizhevsky2009learning}, Street View House Numbers (SVHN) \cite{goodfellow2013multi}, and Food-101 \cite{bossard2014food} are used. CIFAIR-100 has $50,000$ and $10,000$ training and validation images, respectively, with the resolution of $32\times 32$ and $100$ categories; SVHN is a digit classification dataset that has $73,257$ training sample and $26,032$ testing samples; Food-101 includes $101$k images of 101 food categories with each of them has $750$ training and $250$ testing samples. 

\begin{table*}[t]
 \setlength{\tabcolsep}{0.1em}
  \caption{Comparison of Top-1 accuracy using varied amount of parameters adjusted by $d_{bottle}$, different pre-training domains, and the number of frames with other fine-tuning strategies.}
  \label{tab:comparison_petls}
\begin{center}
    \resizebox{0.99\linewidth}{!}{
     {\renewcommand{\arraystretch}{1.2}
  \begin{tabular}{lccccccc}
     \hline
     \multirow{2}[3]{*}{Method} & \multirow{2}[3]{*}{$d_{bottle}$} & \multirow{2}[3]{*}{Pre-training} & \multirow{2}[3]{*}{\# Frames} & \multicolumn{2}{c}{SSv2} & \multicolumn{2}{c}{HMDB51} \\
    \cmidrule(lr){5-6} \cmidrule(lr){7-8}
    & & & & \# Params & Top-1 (\%)  & \# Params & Top-1  (\%)  \\
     \hline
    Full-tuning     & - & Kinetics 400  & 8    & 87.82M  &  \textbf{50.99}  &  87.69M  &  68.07  \\
    Tune FC Layer    & - & Kinetics 400  & 8   & 0.18M  &  24.13  &  0.05M  &  \textbf{71.28}  \\
      LoRA \citep{hu2022lora}   & 16 & Kinetics 400  & 8     & 0.95M  &  38.34  &  0.77M  &  62.12  \\  
    BitFit \citep{zaken2022bitfit}   & - & Kinetics 400  & 8     & 1.29M  &  45.94  &  1.11M  &  68.26  \\
    AdaptFormer-Swin \citep{chen2022adaptformer}   & 64 & Kinetics 400  & 8     & 1.73M  &  40.80  &  1.61M  &  68.66  \\
    Prefix-tuning \citep{li2021prefix}   &  128 & Kinetics 400  & 8     & 6.57M  & 39.46 & 6.40M  & 56.13   \\
      \hline
  Our Swin-BAPAT (w/o Adapter)   & 32 & Kinetics 400  & 8     & 1.35M  &  46.26  &  1.17M  &  69.51  \\   
    Our Swin-BAPAT (w/o Adapter)   & 64 & Kinetics 400  & 8     & 2.51M  &  49.23  &  2.34M  &  \textbf{71.34}  \\   
 Our Swin-BAPAT (w/o Adapter)   & 128 & Kinetics 400  & 8     & 4.83M  &  52.57  &  4.65M  &  70.56  \\  
  Our Swin-BAPAT (w/o Adapter)   & 256 & Kinetics 400  & 8     & 9.45M  & \textbf{52.71}   &  9.27M  &  70.23  \\ 
     \hline
  Our Swin-BAPAT     & 32 & Kinetics 400  & 8     & 2.91M  &  49.63  &  2.74M  &  68.20  \\
  Our  Swin-BAPAT     & 64 & Kinetics 400  & 8     & 4.07M  &  51.80  &  3.89M  &  70.10  \\
 Our  Swin-BAPAT    & 128 & Kinetics 400  & 8      & 6.38M  &  53.36  &  6.20M  &  \textbf{71.93}  \\
 Our  Swin-BAPAT  & 256 & Kinetics 400  & 8   & 11.00M  & \textbf{53.98}  &  10.83M  &  69.64
    \\
    \hline
 Our  Swin-BAPAT  & 256 & Kinetics 400  &  8   & 11.00M  & 53.98  &  10.83M  &  69.64    \\
  Our  Swin-BAPAT  & 256 & Kinetics 600  &  8    & 11.00M  & \textbf{54.06}  &  10.83M  &  \textbf{69.90} \\
    Our  Swin-BAPAT  & 256 & ImageNet-22K  &  8    & 11.00M  &  43.56 &  10.83M  &  59.89  \\
     \hline
 Our   Swin-BAPAT  & 128 & Kinetics 400  & 8    & 6.38M  & 53.36  &  6.20M  &  71.93 \\
 Our   Swin-BAPAT  & 128 & Kinetics 400  & 16    & 6.38M  & \textbf{63.14}  &  6.20M  & \textbf{75.67} \\   
     \hline
  \end{tabular}
  }
  }
  \end{center}
\end{table*}

\textbf{Implementation details}: 
It is worth noting that big batch size (i.e., $1,024$) and the number of input video frames (i.e., 32 frames) can greatly benefit good performance \citep{carreira2017quo, liu2022video, chen2022adaptformer}, which usually requires GPU clusters to enable the training. AdaptFormer \citep{chen2022adaptformer} uses such powerful GPU cluster to achieve good performance. However, good performance might not hold when the batch size is small. Following the more common hardware device setup, we use $4$ GeForce 3090 GPUs for all experiments, leading to a batch size of $64$. All the experiments are fine-tuned for $70$ epochs. We use the Swin-B\footnote{ \url{https://github.com/SwinTransformer/Video-Swin-Transformer}} model pre-trained on Kinetics 400 and 600. For HMDB51, we report the results without tuning the FC layer due to the significant effect of the FC layer on relatively small scale dataset. Following \cite{chen2022adaptformer}, we do not perform regularization strategies such as mixup, cutmix, color jittering, etc. Our PATT module is convenient to be applied to other Transformer-based models. Hence, we respectively adopt ViT-B models from MAE \citep{he2022masked} and VideoMAE \citep{tong2022videomae}  to conduct further comparison on video and image datasets, which follows the self-supervised pre-training setting\footnote{ \url{https://github.com/ShoufaChen/AdaptFormer/blob/main/PRETRAIN.md}} in \cite{chen2022adaptformer} except that the batch size is set to $256$ instead of $1,024$.

\subsubsection{Baselines}
We mainly compare our method Swin-BAPAT with three baselines as follows:\\
(1) \textbf{Full-tuning}: set all the parameters learnable and tune the whole model initiated with the pre-trained weights. \\
(2) \textbf{Tune FC layer}: tune the last fully connected layer and freeze pre-trained parameters of the whole backbone model.\\
(3) \textbf{LoRA}: follow the initial practice \cite{hu2022lora} by adding LoRA matrix to the Q and V of the attention module.\\
(4) \textbf{BitFit}: by tuning the bias of the backbone model together with the FC layer.\\
(5) \textbf{AdaptFormer-Swin}: method introduced by \cite{chen2022adaptformer} that adds a parallel Adapter to the MLP module in each block of the backbone model.\\
(6) \textbf{Prefix-tuning}: the direct implementation of prefix-tuning used in NLP as defined in Eq. \ref{eq:head}. 

\subsection{The Effect of Different PETL Techniques}
Table \ref{tab:comparison_petls} shows the results of different PETL techniques. From the results of four baseline methods, full-tuning performs the best for the large-scale dataset SSv2, whereas tuning the FC layer achieves superior performance over other PETL techniques on HMDB51. This is due to the fact that downstream tasks with relatively larger scale datasets are more parameter hungry for good convergence. On the contrary, small datasets can make good use of the knowledge from the source domain with slight effort of adaption via an FC layer. 
Here, a question regarding the effect of this FC layer when using it together with other PETL techniques has not been investigated. As this FC layer having small amount of tunable parameters can already make a big difference, performing better than full-tuning and other PETL techniques and rendering them not effective for small-scale datasets. As such, we further examine this question in Section \ref{subsec:effect_fc_layer}.

We test different amount of parameters adjusted by $s_{bottle}$, taking its values to $32$, $64$, $128$ and $256$. The second and third groups (without or with Adapter, respectively) of results in Table \ref{tab:comparison_petls} shows that larger values of $s_{bottle}$ can benefit the fine-tuning with slightly more overhead of parameters on large-scale datasets such as SSv2. All results of our Swin-BAPAT outperform the state-of-the-art AdaptFormer-Swin with a big margin (using the smallest value $s_{bottle}=32$ can improve  AdaptFormer-Swin by almost $25\%$). While without using Adapter, our method still outperforms baselines AdaptFormer-Swin and BitFit with roughly similar amount of parameters. When $s_{bottle}$ is larger than $64$, our Swin-BAPAT starts to perform better than full-tuning on both datasets with proper parameter-accuracy trade-off, validating the effectiveness of our Swin-BAPAT for PETL.  

\subsection{The Effect of Different Pre-training Domains}
The knowledge from the pre-trained model is learned from the source domain. We test two different models pre-trained on large-scale datasets: Kinetics 400, Kinetics 600, and ImageNet-22K. Findings show that both two models pre-trained on such large-scale datasets can benefit our proposed PETL strategy with the latter being slightly more significant (see the third group of comparison in Table \ref{tab:comparison_petls}). This is due to the fact that Kinectics 600 is larger than its 400 version and brings more knowledge to the pre-trained model, benefiting more downstream tasks. However, image-based pre-training usually do not perform as good as video-based pre-training due to the larger domain gap.

\begin{figure}
  {\includegraphics[width=\linewidth]{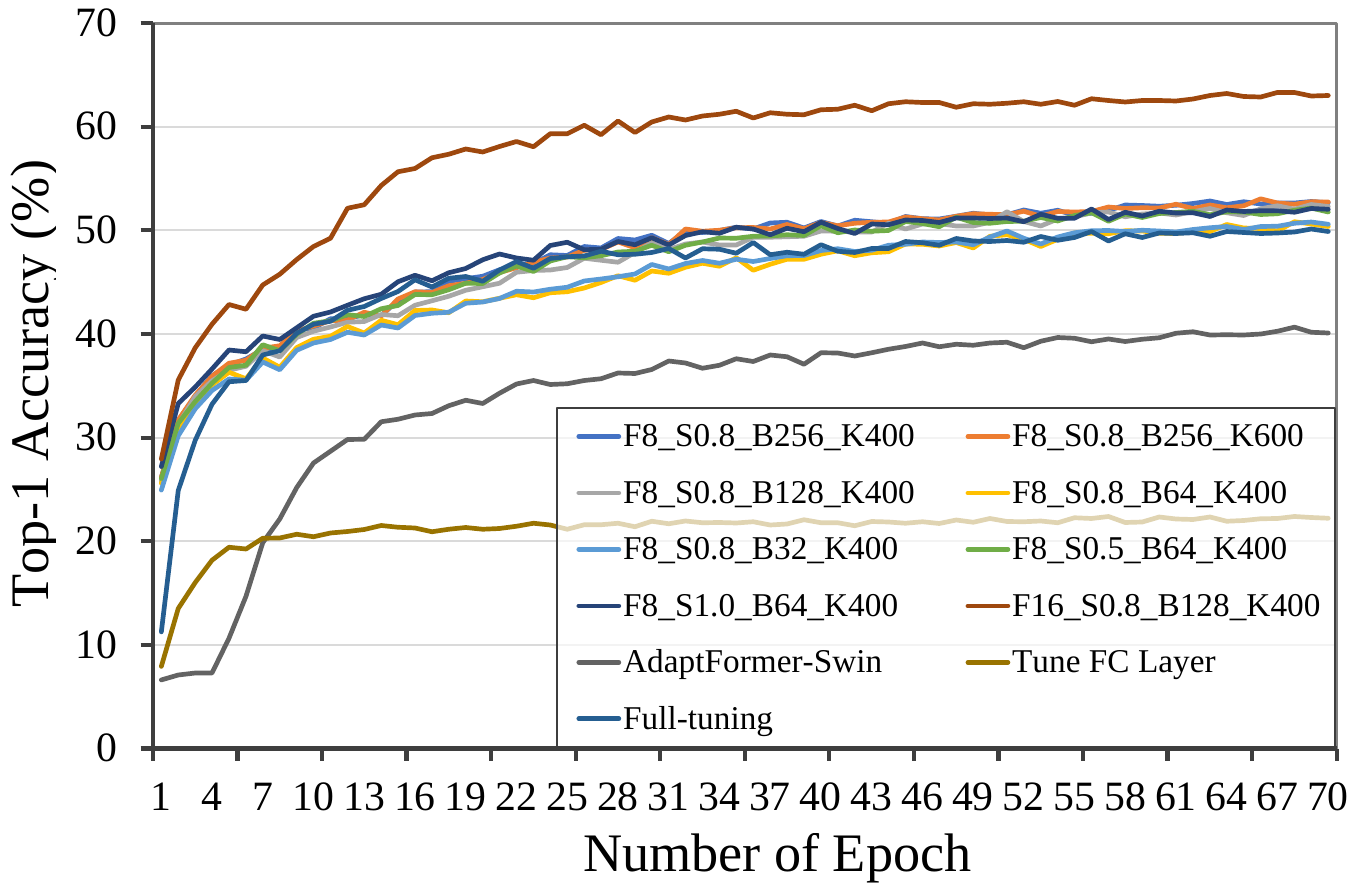}}
  {\caption{Top-1 accuracy of different settings on SSv2 throughout  training process. F: frame, S: scalar, B: $d_{bottle}$, K: pre-training domain.}\label{fig:progress_ssv2}}
\end{figure}

\begin{table*}[h]
  \caption{Results of with or without tuning the FC layer on the small scale dataset HMDB51.}
  \label{tab:comparison_petls_fc}
\begin{center}
    \resizebox{0.95\linewidth}{!}{%
  \begin{tabular}{lccccccc}
    \toprule
     \multirow{2}[3]{*}{Method} & \multirow{2}[3]{*}{$d_{bottle}$} & \multirow{2}[3]{*}{Pre-training} & \multirow{2}[3]{*}{\# Frames} & \multicolumn{2}{c}{with FC layer} & \multicolumn{2}{c}{without FC layer } \\
    \cmidrule(lr){5-6} \cmidrule(lr){7-8}
    & & & & \# Params & Top-1 (\%)  & \# Params & Top-1  (\%)  \\
    \midrule
  Our  Swin-BAPAT     & 32 & Kinetics 400  & 8     & 2.79M  &  65.97  &  2.74M  &  68.20  \\
  Our   Swin-BAPAT     & 64 & Kinetics 400  & 8     & 3.94M  &  67.28  &  3.89M  &  70.10  \\
  Our  Swin-BAPAT    & 128 & Kinetics 400  & 8      & 6.25M  &  66.75  &  6.20M  &  \textbf{71.93}  \\
  Our  Swin-BAPAT  & 256 & Kinetics 400  & 8   & 10.88M  & \textbf{67.67}  &  10.83M  &  69.64
    \\
    \midrule
  Our  Swin-BAPAT  & 256 & Kinetics 400  &  8   & 10.88M  & \textbf{67.67}  &  10.83M  &  69.64    \\
  Our   Swin-BAPAT  & 256 & Kinetics 600  &  8    & 10.88M  & 67.41  &  10.83M  &  \textbf{69.90} \\
    \midrule
   Our   Swin-BAPAT  & 128 & Kinetics 400  & 8    & 6.25M  & 66.75  &  6.20M  &  71.93 \\
   Our  Swin-BAPAT  & 128 & Kinetics 400  & 16    & 6.25M  & 70.56  &  6.20M  & 75.67 \\
  Our   Swin-BAPAT  & 128 & Kinetics 400  & 32    & 6.25M  & \textbf{74.82}  &  6.20M  &  \textbf{76.46} \\    
    \bottomrule
  \end{tabular}
  }
  \end{center}
\end{table*}

\begin{table}[t]
  \caption{Top-1 accuracy (\%)  using different scalar values on two datasets: SSv2 and HMDB51. The $d_{bottle}$ is set to $128$; pre-training is based on Kinetics 400.}\label{tab:scalar}
  \begin{center}
    \resizebox{0.9\linewidth}{!}{
     {\renewcommand{\arraystretch}{1.2}
    \begin{tabular}{lcc}
        \toprule
      \multicolumn{1}{l}{\bf Scalar $s$}  &\multicolumn{1}{c}{\bf SSv2} &\multicolumn{1}{c}{\bf HMDB51} \\
    \midrule
        Full-tuning  &  50.99  & 71.28 \\
        Tune FC Layer  &  24.13  & 68.07 \\
        AdaptFormer-Swin &  40.80  & 68.66 \\
     \midrule
        $s=0.2$  &  47.46  & 69.38 \\
        $s=0.5$  &  52.84  & 71.87 \\
        $s=0.8$  &  \textbf{53.36}  & \textbf{71.93} \\
        $s=1.0$  &  53.29  & 70.89 \\
        \bottomrule
    \end{tabular}
    }
    }
    \end{center}
\end{table}
\subsection{The Effect of Different Video Input Size}
We also test whether our method is robust to increased number of input video frames. It is worth noting that larger number of input video frames usually can bring more spatial temporal information, benefiting data-driven models to learn more distinguishable features while keeping the model size remaining the same. The last group of comparisons in Table \ref{tab:comparison_petls} shows that using double-sized video input (i.e., $16$ frames) can greatly improve the performance of action recognition on both small and large-scale datasets. The improvements (increased  $9.78\%$ from $53.36\%$ to $63.14\%$ on SSv2, and $3.74\%$  from $71.93\%$ to $75.67\%$ on HMDB51) are more significant than other factors such as $d_{bottle}$ and pre-training domain (around $1\%$ to $2\%$). The top line in Figure \ref{fig:progress_ssv2} visualizes the significant effect of increasing the number of input video frames. These results suggest that our Swin-BAPAT can be promising for increased frames of video input.

\subsection{The Effect of Different Scale of PATT}
Recall that the effect of our PATT on pre-trained models can be adjusted by the variable $s$ in Eq. \ref{eq:head2}. Table \ref{tab:scalar} shows that adopting the value of $0.8$ can deliver consistent best performances on both datasets SSv2 and HMDB51 under our experimental setting. Smaller values of $s$ will quantitatively reduce the effect of our PATT module on the knowledge transfer while large values will increase the effect of our PATT module. The good performance achieved via taking an effective scale of $0.8$ indicates that our PATT module plays an important role in the knowledge transfer. However, even larger values over $0.8$ can affect the importance of original knowledge thereof the pre-trained model. Hence, proper valued scalar $s$ is essential for balancing the role of PATT and pre-trained backbone model. Note this can be a learnable parameter upon specific implementation, here we empirically verified the effect of the scalar.

\begin{table}[t]
  \caption{Ablation of different implementation positions of PATT defined in Eq. \ref{eq:head2}, e.g., Ours ($\displaystyle \tK$, $\displaystyle \tV$) indicates inserting PATT to the query and key of 3DSW-MSA modules. Pre-training on Kinetics 600. $d_{bottle}$ is set to 128; Scalar $s$ is set to $0.8$.}
  \label{tab:comparison_petls_qkv_combinations}
\begin{center}
    \resizebox{\linewidth}{!}{%
    {\renewcommand{\arraystretch}{1.2}
  \begin{tabular}{lcccc}
    \toprule
     \multirow{2}[3]{*}{Method} & \multicolumn{2}{c}{SSv2} & \multicolumn{2}{c}{HMDB51} \\
    \cmidrule(lr){2-3} \cmidrule(lr){4-5}
    & \# Params & Top-1 & \# Params & Top-1  \\
    \hline
    Full-tuning       & 87.82M  &  50.99  &  87.69M  &  68.07  \\
 \hline
 Concat ($\displaystyle \tK$, $\displaystyle \tV$)       & 6.38M  &  15.61  &  6.20M  &  20.98  \\     
 No $Z^{l-1}$  ($\displaystyle \tK$, $\displaystyle \tV$)      & 8.74M  &  51.06  &  8.56M  &  67.41  \\      
    \hline
    Ours ($\displaystyle \tQ$, $\displaystyle \tK$)         & 6.38M  &  45.49  &  6.20M  &  68.92  \\
     Ours ($\displaystyle \tK$, $\displaystyle \tV$)       & 6.38M  &  \textbf{53.38}  &  6.20M  &  71.41  \\
      Ours ($\displaystyle \tQ$, $\displaystyle \tV$)     & 6.38M  &  53.24  &  6.20M  &  \textbf{71.74}  \\
      Ours ($\displaystyle \tQ$, $\displaystyle \tK$, $\displaystyle \tV$)   & 7.93M  &  53.23  &  7.63M  &  69.57  \\
    \hline
  \end{tabular}
  }
  }
  \end{center}
\end{table}
\begin{table*}[t]
  
  \caption{Comparison of Top-1 accuracy via ViT-B models with supervised pre-training for image and video tasks. * indicates reproduction via our pre-training weights.}
  \label{tab:comparison_petls_video_image_supervised}
\begin{center}
    \resizebox{\linewidth}{!}{%
    \setlength{\tabcolsep}{0.3em}
    {\renewcommand{\arraystretch}{1.2}
  \begin{tabular}{l|c|ccc|cc}
    \hline
     \multirow{2}[3]{*}{Method\vspace{5pt}} & Avg. 
     & \multicolumn{3}{c|}{Image} & \multicolumn{2}{c}{Video} \\
    & Params (M) & CIFAR-100 & SVHN  &  Food-101  & SSv2 & HMDB51  \\
\hline
\rowcolor{grayrow}
Full-tuning *  &  86.04 (100\%)  &  90.00  &  97.54  &  89.70  &  51.36  &  57.51   \\
Tune FC layer *  &  0.07 (0.08\%)  &  60.44\space \color{blue} (-29.56)  &  60.44\space \color{blue} (-37.10)  &  54.07\space \color{blue} (-35.63)  &  37.52\space \color{blue} (-13.84)  &  68.07\space \color{red} (+10.56)   \\
AdaptFormer-64 *  &  1.26 (1.46\%)  &  90.80\space \color{red} (+0.80)  &  96.73\space \color{blue} \color{blue} (-0.81)  &  89.15\space \color{blue} (-0.55)  &  54.70\space \color{red} (+3.34)  &  66.03\space \color{red} (+8.52)   \\
\hline
Our ViT-BAPAT-32  &  2.13 (2.47\%)  &  91.79\space \color{red} (+1.79)  &  96.97\space \color{blue} (-0.57)  &  89.43\space \color{blue} (-0.27)  &  59.49\space \color{red} (+8.13)  &  73.31\space \color{red} (+15.80)   \\
Our ViT-BAPAT-64  &  3.02 (3.51\%)  &  91.55\space \color{red} (+1.55)  &  96.91\space \color{blue} (-0.63)  &  89.63\space \color{blue} (-0.07)  &  59.32\space \color{red} (+7.96)  &  73.38\space \color{red} (+15.87)   \\
Our ViT-BAPAT-128  &  4.79 (5.56\%)  &  91.45\space \color{red} (+1.45)  &  97.05\space \color{blue} (-0.49)  &  89.48\space \color{blue} (-0.22)  &  58.70\space \color{red} (+7.34)  &  73.83\space \color{red} (+16.32)   \\
Our ViT-BAPAT-256  &  8.33 (9.68\%)  &  91.57\space \color{red} (+1.57)  &  96.88\space \color{blue} (-0.66)  &  89.43\space \color{blue} (-0.27)  &  57.58\space \color{red} (+6.22)  &  73.97\space \color{red} (+16.46)   \\
\hline
  \end{tabular}
  }
  }
  \end{center}
\end{table*}

\subsection{The Effect of Different Methods yield from V-PETL}
We have argued that, especially for relative large downstream datasets, the position and the amount of trainable parameters are important for parameter-efficient transfer learning in Section \ref{subsec:v_petl}. The proposed Swin-BAPAT is one of instantiated models from the V-PETL framework regarding the insert position of our PATT. Other instantiations can be inserted into different positions such as query, key, and value of the attention module. We further instantiate other variations of our Swin-BAPAT by inserting PATT to different positions. Table \ref{tab:comparison_petls_qkv_combinations} shows the results. Findings show that inserting to the value position of 3DSW-MSA can contribute more than inserting to other two positions. While inserting to query of key makes little difference for the performance. This is due to the fact that query and key make the calculation of the attention mask. Hence, inserting either one of them will lead to a similar effect. 
On one hand, these results, to some extent, justify the original design of prefix-tuning that bring learnable prefix to key and value of the attention module. On the other hand, it indicates that our claim regarding the unified view of PETL for visual tasks is reasonable. In Table \ref{tab:comparison_petls_qkv_combinations}, we also ablate the designs of PATT regarding concatenating $K_p$ and $V_p$ (i.e.,  Concat [$\displaystyle \tK$, $\displaystyle \tV$]), and using trainable parameters to generate $K_p$ and $V_p$ (i.e.,  No $Z^{l-1}$  [$\displaystyle \tK$, $\displaystyle \tV$]).

\subsection{The Effect of FC Layer for Small Scale Downstream Tasks}
\label{subsec:effect_fc_layer}

For the small dataset HMDB51, due to the good parameter-accuracy trade-off achieved by fine-tuning the FC layer only, adding the FC layer cannot bring extra improvement to our proposed method. Without sufficient training data, full-tuning also cannot perform well (see results in Table \ref{tab:comparison_petls}). As such, small datasets do not need to rely on LVMs but can make use of LVMs with light transfer. Instead, without tuning the FC layer, our Swin-BAPAT can perform better than fine-tuning the FC layer with small amount of extra trainable parameters (see results in Table \ref{tab:comparison_petls_fc}), validating the good parameter-accuracy trade-off of our method.

\subsection{Results with More Supervised Pre-trained Models on More Downstream Vision Tasks}
We conduct further experiments for image and video tasks using weights of supervised pre-training. For image tasks, we use the pre-trained weights on ImageNet\_21K vit\_base\_patch16\_224\_miil\_in21k\footnote{ \url{https://miil-public-eu.oss-eu-central-1.aliyuncs.com/model-zoo/ImageNet_21K_P/models/timm/vit_base_patch16_224_in21k_miil.pth}}. For video tasks, we use ViT-B of VideoMAE fine-tuned on Kinetics-400 (Top-1=$81.5\%$)\footnote{ \url{https://drive.google.com/file/d/1MzwteHH-1yuMnFb8vRBQDvngV1Zl-d3z/view?usp=sharing}}. Since AdaptFormer has not provided the supervised pre-trained weights, our supervised pre-training settings might differ from the ones used by AdaptFormer. Hence, we also conduct experiments for full-tuning, linear probing, and AdapFormer-64 with our experimental settings (i.e., smaller batch size). The results are shown in Table \ref{tab:comparison_petls_video_image_supervised}. Throughout five vision tasks, our method ViT-BAPAT-32 consistently outperforms the state-of-the-art method AdaptFormer \citep{chen2022adaptformer} with a comparable amount of trainable parameters, validating the effectiveness of our PATT module. On video-based tasks, our method even achieved more significant improvement than that of Video Swin Transformer in Table \ref{tab:comparison_petls} over full-tuning and linear probing (i.e., tune FC layer). While for image-based tasks, the performance can be  competitive with full-tuning.

\begin{table*}[h]
  
  \caption{Comparison of Top-1 accuracy via ViT-B models from MAE and VideoMAE pre-trained with self-supervised learning for image and video datasets, respectively.}
  \label{tab:comparison_petls_video_image}
\begin{center}
    \resizebox{\linewidth}{!}{
    \setlength{\tabcolsep}{0.3em}
    {\renewcommand{\arraystretch}{1.2}
  \begin{tabular}{l|c|ccc|cc}
    \hline
     \multirow{2}[3]{*}{Method\vspace{5pt}} & Avg.
     & \multicolumn{3}{c|}{Image} & \multicolumn{2}{c}{Video} \\
    & Params (M) & CIFAR-100 & SVHN  &  Food-101  & SSv2 & HMDB51  \\
   \hline
 \rowcolor{grayrow}
Full-tuning  &  86.04 (100\%)  &  85.90  &  97.67  &  90.09  &  53.97  &  46.41   \\
Tune FC Layer  &  0.07 (0.08\%)  &  69.83\space \color{blue}(-16.07)  &  66.91\space \color{blue}(-30.76)  &  69.74\space \color{blue}(-20.35)  &  29.23\space \color{blue}(-24.74)  &  49.84\space \color{red}(+3.43)   \\
VPT \citep{jia2022visual}  &  0.08 (0.09\%)  &  82.44\space \color{blue}(-3.46)  &  94.02\space \color{blue}(-3.65)  &  82.98\space \color{blue}(-7.11)  &  43.73\space \color{blue}(-10.24)  &  52.67\space \color{red}(+6.26)   \\
AdaptFormer-64 
&  1.26 (1.46\%)  &  85.90\space \color{red}(0.00)  &  96.89\space \color{blue}(-0.78)  &  87.61\space \color{blue}(-2.48)  &  59.02\space \color{red}(+5.05)  &  55.69\space \color{red}(+9.28)   \\
\hline
Our ViT-BAPAT-32  &  2.13 (2.47\%)  &  86.29\space \color{red}(+0.39)  &  97.18\space \color{blue}(-0.49)  &  87.37\space \color{blue}(-2.72)  &  57.78\space \color{red}(+3.81)  &  57.18\space \color{red}(+10.77)   \\
Our ViT-BAPAT-64  &  3.02 (3.51\%)  &  86.35\space \color{red}(+0.45)  &  97.18\space \color{blue}(-0.49)  &  87.53\space \color{blue}(-2.56)  &  57.55\space \color{red}(+3.58)  &  57.18\space \color{red}(+10.77)   \\
Our ViT-BAPAT-128  &  4.79 (5.56\%)  &  86.47\space \color{red}(+0.57)  &  97.28\space \color{blue}(-0.39)  &  87.66\space \color{blue}(-2.43)  &  56.97\space \color{red}(+3.00)  &  57.70\space \color{red}(+11.29)   \\
Our ViT-BAPAT-256  &  8.33 (9.68\%)  &  86.55\space \color{red}(+0.65)  &  97.24\space \color{blue}(-0.43)  &  87.68\space \color{blue}(-2.41)  &  56.53\space \color{red}(+2.56)  &  57.31\space \color{red}(+10.90)   \\
\hline
  \end{tabular}
  }
  }
  \end{center}
\end{table*}

\subsection{Comparison on Varied tasks via Self-supervised Pre-trained Models}
Table \ref{tab:comparison_petls_video_image} shows the comparison with AdaptFormer-64 \citep{chen2022adaptformer} and VPT \citep{jia2022visual} on both image- and video-based downstream tasks. Our method ViT-BAPAT still shows promising parameter-accuracy trade-off via much smaller batch size, which is more convenient for reproduction on the general single server with 8 GPUs. The underperformance on SSv2 (better than full-tuning) can be due to the smaller batch size as SSv2 is much larger than other compared datasets and can be more relying on larger batch size. In real-world application scenarios, small dataset can be the more common case, which confirms our contributions.

\section{Conclusion}
In this paper, we introduced a V-PETL framework for exploiting good  parameter-accuracy trade-off around adapting LVMs to downstream tasks. Our Swin-BAPAT method derived from the V-PETL with a variation of prefix-tuning known as PATT can effectively bring good parameter-accuracy trade-off on downstream tasks. The proposed PATT can be easily plugged to the attention module of other transformer-like models. Meanwhile, the amount of trainable parameter can be easily adjusted by the parameter $d_{bottle}$. With small amount overhead on trainable parameters, our method performs significantly better than state-of-the-art method AdapFormer-Swin and full-tuning on the datasets SSv2 and HMDB51 via small batch size, validating our contribution to the literature of PETL. Further extensive experiments on more downstream tasks with more LVMs have also verified the effectiveness of our PATT module. 

In the future we will test our proposed model on more action recognition datasets surveyed in \cite{sun2022human} under more learning regimes such as zero/few-shot learning, active learning, and continual learning with other pre-training methods such as visual-language models. We will also explore other backbone models, activation functions for PATT, and PETL techniques such as Compacter \citep{karimi2021compacter},  ST-Adapter, and KAdaptation for visual tasks.

\backmatter
\bmhead{Acknowledgments}
This work is supported in part by Huawei Technologies under Grant No.: P0038941. The author would like to thank the editors and anonymous reviewers who help improve this manuscript.

{
\setlength{\bibsep}{0pt plus 0.3ex}
\footnotesize
\bibliography{references}

}


\end{document}